\documentclass[sigconf]{acmart}
 \setcopyright{none}
\renewcommand\footnotetextcopyrightpermission[1]{}

\usepackage{pgfplots}
\pgfplotsset{compat=1.18}
\usepackage{tikz}
\usepackage{pgf-pie}
\usepackage{hyperref}

\AtBeginDocument{%
  }

\copyrightyear{2026}
\acmYear{2026}
\setcopyright{none}

\fancypagestyle{customheader}{
  \fancyhf{}
  \fancyhead[C]{%
    \small Accepted in 2026 15th ICSCA, Langkawi, Malaysia\\
  }
}

\begin{document}
\thispagestyle{firstpagestyle}
\title{Addressing Data Scarcity in Bangla Fake News Detection: An LLM-Based Dataset Augmentation Approach}

\author{Ahmed Alfey Sani}
\affiliation{%
  \institution{Islamic University of Technology}
  \city{Gazipur}
  \country{Bangladesh}
  }
\email{ahmedalfey@iut-dhaka.edu}

\author{Kazi Akib Zaoad}
\affiliation{%
  \institution{Islamic University of Technology}
  \city{Gazipur}
  \country{Bangladesh}
}
\email{akibzaoad@iut-dhaka.edu}

\author{Shefayat E Shams Adib}
\affiliation{%
  \institution{Islamic University of Technology}
  \city{Gazipur}
  \country{Bangladesh}
}
\email{shefayatadib@iut-dhaka.edu}

\author{Md Abdul Muqtadir}
\affiliation{%
  \institution{Islamic University of Technology}
  \city{Gazipur}
  \country{Bangladesh}
}
\email{abdulmuqtadir@iut-dhaka.edu}

\author{Ajwad Abrar}
\affiliation{%
  \institution{Islamic University of Technology}
  \city{Gazipur}
  \country{Bangladesh}
}
\email{ajwadabrar@iut-dhaka.edu}

\renewcommand{\shortauthors}{Sani et al.}

\begin{abstract}
 The growing spread of misinformation in digital media highlights the need for reliable fake news detection systems, yet progress in under-resourced languages such as Bangla is limited by small and imbalanced datasets. This study investigates whether Large Language Model (LLM)–based augmentation can effectively address this limitation and improve Bangla fake news classification. Existing datasets remain valuable but highly imbalanced, limiting model performance, and LLM-based augmentation for Bangla has been scarcely explored. To fill this gap, we propose a systematic augmentation framework that generates synthetic Bangla news articles using the instruction-tuned Gemma-3-27B-IT model, supported by semantic filtering and controlled subsampling to preserve label consistency and diversity. We compare zero-shot and few-shot prompting, evaluate multiple augmentation rates, and examine random versus similarity-based selection strategies. Our experiments show that augmenting only the minority class with a high augmentation rate and random subsampling yields the strongest gains, raising the Fake News F1 score from 0.85 to 0.88. To support reproducibility and further research in this low-resource domain, we publicly release 4,545 synthetically generated Bangla fake news samples along with our full implementation. These findings demonstrate that well-designed LLM-driven augmentation can significantly improve fake news detection in low-resource settings and provide a practical foundation for advancing multilingual misinformation research.

\end{abstract}



\maketitle
\thispagestyle{customheader}

\section{Introduction}

Online platform emergence has transformed news consumption, being the most powerful information provider to a networked global population. In recent years, digital entitlement news has emerged as an indictment of all media. This has been seen in coverage of journalist layoffs and decreasing circulation \cite{staubin2024news}. Research indicates that 86\% of adults use digital means to access news. They either do this frequently or sometimes they rely on news through social media. News consumption surges in times of high information intensity, for example during extraordinary or unpredictable events, which implies exposure to disinformation will be higher. Examples of these events would be the U.S. presidential elections, COVID-19 pandemic and Russia–Ukraine war \cite{grinberg2019fake, gabarron2021covid, sanchez2024disinformation, pierri2023propaganda}.

Artificial Intelligence (AI) has been developing rapidly and makes the battle against disinformation even more complex. While the technology of generative models as a resource for misinformation can be useful in fact-checking, the available evidence suggests it would operate as enablers of fake news \cite{feuerriegel2024generative, feuerriegel2023research}. Although regarded as less reliable, misinformation produced by AI machines can have widespread reach and impact due to its expansive scale, rapidness and access \cite{feuerriegel2023research, bashardoust2024comparing}. Fake news has been correlated, in terms of cause or effect, with increasing rates of psychological and physical disorders (such as anxiety and obesity), rising skepticism about science, and erosion of democracy and freedom of speech \cite{rocha2023impact, alonso2020impact}. Research has shown that public trust in information and news is significantly shaped by the presence of fact-checking mechanisms on social media platforms \cite{olan2024fake}. These findings highlight the need for stronger and more widely available fact-checking services within social media ecosystems, along with improved support for news verification.

There is a huge volume of work focused on this fundamental problem of fake news detection. The paper on fake news in 2016 US Presidential Election by Allcott et al. \cite{allcott2017social} is the source of a widely used definition for "fake news" in terms of news stories which are "intentionally and verifiably false, and could mislead readers". Alghamdi et al. \cite{alghamdi2024comprehensive} offers a comprehensive survey of fake news research and detection techniques, including state-of-the-art machine learning methods, such as deep learning. The quality of training data plays a key role in both the creation and accuracy of supervised models. However, a significant bottleneck for fake news detection is the lack of enough high-quality labeled datasets especially for under-resourced languages.

Although several datasets have been developed for high-resource languages, resources are limited for low-resource languages such as Bangla. Hossain et al. \cite{hossain2020banfakenews} attempts to fill in this gap, by proposing \textbf{BanFakeNews}\footnote{\url{https://www.kaggle.com/datasets/cryptexcode/banfakenews}} – a curated Bangla news dataset for the task of fake news detection. Despite this stride, a number of these datasets have their own issues including small sample size, class imbalance, lack of fine granularity and being prone to obsoleteness as fake news progresses \cite{hamed2023review, kuntur2024fake}. Especially class imbalance often results in biased models whose performance is biased towards the majority class, which is usually real news for fake news detection tasks. This problem is common in most published fake news datasets, however relatively few research works aim to solve the data imbalance \cite{alnabhan2024fake}.

In response to this, researchers have investigated methods to augment the existing fake news datasets in order to increase the volume and variety of training data. Common techniques are paraphrasing, synonym replacement and back-translation, realized on rule based constraints systems or more sophisticated algorithms \cite{amjad2020data, kapusta2024text, junior2022use, hua2023multimodal, keya2022augfake, ashraf2021cic}. Large Language Models (LLMs) have transformed Natural Language Processing (NLP) \cite{zhao2023survey, minaee2024large}. Their ability to generate high-quality, human-like synthetic text makes them a promising tool for data augmentation, offering a way to enrich and expand human-annotated datasets. Although LLMs have shown strong performance in data augmentation across various NLP tasks \cite{ding2024data, hu2024bad}, their use in non-English fake news detection remains underexplored.

We address the scarcity and imbalance of large-scale Bangla fake news datasets by leveraging an open-source LLM from Google DeepMind to generate synthetic BanFakeNews samples. To guide the model, we design targeted prompts that produce multiple synthetic variants of a given news article while preserving its real or fake label. Building on prior work in multilingual zero-shot prompting \cite{chalehchaleh2025enhancing}, we extend the investigation to a few-shot setting, where a small number of example articles are included in the prompt to condition the generation process. This enables a direct comparison between zero-shot and few-shot augmentation in terms of both sample quality and downstream detection performance.

We further explore key factors that influence augmentation effectiveness, including different augmentation rates, random and similarity-based subsampling strategies, and class-centric augmentation. The augmented datasets are then used to train BERT-based fake news classifiers \cite{devlin2018bert}. The main contributions of this study are summarized as follows:

\begin{itemize}
    \item We introduce a data augmentation framework using a large language model (Gemma-3-27B-IT) to address the low-resource challenge in Bangla fake news detection, demonstrating how synthetic samples can enhance performance in underrepresented languages.
    
    \item We conduct a systematic evaluation of augmentation strategies, comparing prompting methods (zero-shot and few-shot), augmentation rates (K=1, 2, 3), and sample selection techniques (random and similarity-based) to identify the most effective configuration for improving detection accuracy.
    
    \item We release 4,545 synthetically generated Bangla fake news samples as an extension of the BanFakeNews dataset, providing a valuable resource for future research. The full implementation is publicly available\footnote{\url{https://github.com/phigratio/bangla-fake-news}} for reproducibility and to support future research.

\end{itemize}

In Sections~\ref{sec:lit-rev} and~\ref{sec:methodology}, we reviewed the related work and outlined the motivation behind our augmentation techniques and classification approach. Section~\ref{sec:experiments} presents the experimental setup using the BanFakeNews dataset, followed by Section~\ref{sec:results}, which reports the results across different augmentation configurations. Sections~\ref{sec:discussion} and~\ref{sec:conclusion} conclude the paper with a discussion of the implications, limitations, and potential directions for future research.

\section{Literature Review} 
\label{sec:lit-rev}
\subsection{Fake News Detection: Overview and Challenges}
The rise of online platforms has reshaped news consumption, with digital media now serving as the primary information source for an increasingly connected global population. During high-impact events such as elections, pandemics, and geopolitical crises, news engagement intensifies, increasing public exposure to misinformation \cite{grinberg2019fake, gabarron2021covid}. The rise of generative artificial intelligence has further intensified these challenges, with AI-generated misinformation achieving broad reach and influence despite being perceived as less accurate \cite{bashardoust2024comparing}. Fake news is connected to mental and physical health problems, increasing mistrust in science and the decline of democracy and freedom of expression \cite{rocha2023impact, alonso2020impact}. Lots of research has been conducted on the significant problem for detecting the fake news. Allcott et al. \cite{allcott2017social} provided a widely cited definition of "fake news," describing it as "news articles that are intentionally and verifiably false, and could mislead readers." Alghamdi et al. \cite{alghamdi2024comprehensive} provided a comprehensive review of fake news detection methods, covering machine learning and deep learning approaches that use content, context, and hybrid features. A key challenge in this area is the shortage of large, high-quality labeled datasets, particularly for under-resourced languages. The dynamic nature of fake news, along with persistent class imbalance, further complicates the development of robust detection systems.

\subsection{Data Limitations in Fake News Detection}
Adequate availability of high-quality data is essential for training robust and reliable machine learning models. The efficiency and the accuracy of such real-word models are intrinsically linked to the datasets on which they operate to train them. Although many fake news datasets exist for fake news detection \cite{d2021fake}, they often suffer from the drawbacks of such small size, class imbalance, insufficient granularity, limited language coverage, and the risk of becoming outdated due to the dynamic nature of fake news \cite{hamed2023review,kuntur2024fake}. These problems are even more exacerbated for under-resourced languages. Also, focusing on the class imbalance problem, Alnabhan et al. \cite{alnabhan2024fake} offered a review of deep learning strategies of fake news detection. The lack of real news samples and the abundance of fake news samples can result biased models, which performs well on majority class (real news), but not well minority class (fake news). Their review shows that many existing fake news datasets suffer from class imbalance, and only a few studies have directly addressed this issue. The oversampling techniques examined in prior work include random and advanced over- and downsampling methods, as well as the use of alternative loss functions.

\subsection{Fake News Detection in Low-Resource Languages}

Although there is a handful of efforts that have surfaced that focus on fake news in English detection, low resource languages are largely understudied. For Indonesian, Pratiwi et al.  \cite{pratiwi2017study} generated a dataset of 250 articles for hoaxes as well as real news, and developed a naive-Bayes-based hoax detection. For Bangla language, which has around 341 million speakers worldwide and is the sixth-most spoken language in the world \cite{shahriar2024improving}, research on this domain is particularly scarce.

Hossain et al. \cite{hossain2020banfakenews} presented the BanFakeNews dataset, the first of its kind dataset for Bangla while tackling fake news. To build this dataset, the authors collected authentic news from 22 popular and trusted Bangladeshi news portals. Fake news samples included misleading or false-context content, clickbait articles, and satire or parody materials intended for entertainment. The final dataset contains approximately 50{,}000 news articles, enriched with metadata such as domain, publication time, category, source, and headline–article relations.

The BanFakeNews analysis considered the traditional machine learning and neural network paradigms. Traditional models included Support Vector Machine (SVM), Random Forest (RF), and Logistic Regression (LR), using the following linguistic and metadata features:

\begin{itemize}
    \item \textbf{Lexical:} word n-grams (1–3) and character n-grams (3–5) with TF-IDF weighting;
    \item \textbf{Syntactic:} normalized frequencies of POS tags;
    \item \textbf{Semantic:} pre-trained FastText and Word2Vec embeddings;
    \item \textbf{Metadata and punctuation:} punctuation counts, headline/body lengths, and Alexa rankings of publishing sites.
\end{itemize}

Interestingly, linear models using linguistic features outperformed neural models. Character-level features, especially character 3-grams, proved particularly effective. The best-performing systems based on these features surpassed the neural baselines by a notable margin. Human baselines conducted by Hossain et al. \cite{hossain2020banfakenews} yielded interesting observations. Five of the annotators obtained F1-scores between 58\% and 70\% for detecting fakes with an inter-annotator agreement (Fleiss' Kappa) of 38.83\%. They concluded that content credibility and source credibility are the two most important dimensions that humans use to differentiate fake from real news. They also found it difficult to detect fake news when the information seemed convincing and was supported by strong evidence or statistics. These results indicate that the detection performance could be improved and show the significance of source information as a factor.

\subsection{Data Augmentation in Fake News Detection}
Data augmentation applies transformations to existing samples to generate synthetic data, increasing the size and diversity of a training set without requiring new data \cite{bayer2022survey, wang2020generalizing}. Although previous work has investigated how to apply data augmentation to fake news detection, it is still unclear about the effect. Some studies provide evidence of improvements in performance, whilst others show no effect, detrimental effects, or inconsistent effects on detection results. Where effective, conventional forms of augmentation have been inconsistent. Amjad et al. \cite{amjad2020data} used machine translation based augmentation for Urdu fake news classification both manually annotated Urdu text and machine-translated English fake news, but it did not show noteworthy improvement. Kapusta et al. \cite{kapusta2024text} investigated the impact of synonym replacement, back translation and function word reduction for optimizing a corpus for fake news classification and developed several augmented versions to train Word2Vec skip-gram models. They found significant differences in classifiers’ performance between enhancing and original data. But Ashraf et al. \cite{ashraf2021cic} noticed that adding to their news claim dataset with inserted or replaced words using Word2Vec-based similarity search yielded no performance improvements.

\subsubsection{Pre-trained Language Model (PLM)-Based Approaches}
\mbox{}\\
In addition to typical augmentation techniques, there has been recent interest in Pre-trained Language Models (PLMs). The introduction of transformer-based PLMs \cite{vaswani2017attention} represented a paradigm shift in natural language processing, enabling deep, context-aware language comprehension and in resulting the broad dissemination to a variety of NLP tasks. Among those are BERT (Bidirectional Encoder Representations from Transformers) \cite{devlin2019bert} and, its descendent, RoBERTa (Robustly Optimized BERT Approach) \cite{liu2019roberta}, which is more strongly optimized for pre-training than BERT matures. These models model intricate linguistic structures and semantic links by learned from large-scale corpus. Researches are focused to incorporate PLMs in the augmentation pipeline for fake news classification. Hua et al. \cite{hua2023multimodal} introduced a multimodal method that integrates BERT-based back-translation as well as image parsing and contrastive learning. They increase both the dataset size by back-translation and belief refinement of multimodal representations with contrastive learning, achieving promising gains as shown in their experiments. Keya et al. \cite{keya2022augfake} proposed AugFake-BERT, a training strategy that integrates data augmentation with transfer learning to improve fake news datasets. The method uses BERT to synthesize new texts and is able to magnify the dataset using augmented fake data. Their experiments showed that oversampling the minority class resulted in a better classification. On the other hand, Júnior et al. \cite{junior2022use} achieved opposing results when enriching an English dataset by part-of-speech tagging to detect nouns and replaced them with semantically substituted synonyms. The increased dataset was translated to Portuguese and classified by BERT, which performed poorly in validation and even worse in the test sets. This indicates that the success of PLM-based augmentation strategies varies depending on how they are implemented and used.

\subsubsection{Large Language Model (LLM)-Based Approaches}
\mbox{}\\
Even more recently, LLMs have gained substantial popularity in academia, industry and the public press. Such large models, which are built from massive pre-trained neural networks trained on very large corpora, have excelled across various benchmarks. OpenAI’s recent ChatGPT release has reignited interest in LLMs. LLMs perform well across a variety of general-purpose and domain-specific NLP tasks due to their strong natural-language understanding and generation capabilities, which have resulted from being trained on a large number of parameters over an enormous amount of data. Minaee et al. \cite{minaee2024large} provided a comprehensive overview of leading LLMs, highlighting their features, advancements, and limitations. Ding et al. \cite{ding2024data} performed an extensive investigation of the use of LLMs for data augmentation in NLP. Their study investigates the adversarial threats and promise in this approach, exploring a family of strategies that rely on LLMs from the viewpoint of data-centric inference and learningcentric inference. While there is increasing interest in LLMs for augmenting data across multiple NLP tasks, the use of LLMs on fake news detection has yet to be fully explored. The two have just started to be studied in cooperation. Ahlbäck et al. \cite{ahlback2023can} experimented with several data augmentation methods, including LLM-based ones like Vicuna and ChatGPT. They investigated these methods in combination with BERT for fake news detection. There was a non-statistical significant improvement in the results. Wu et al. \cite{wu2024towards} proposed a model-agnostic approach to enhance semantic perception in evidence-aware fake news detection.
It uses two data augmentation strategies: semantic-flipped augmentation (producing opposite-meaning claims via SpaCy and ChatGPT) and semantic-invariant augmentation (paraphrasing with unchanged meaning using PEGASUS or ChatGPT). Together with a claim representation learning module, these augmentations help the model to better identify semantic changes. Although previous work has shown that LLMs have the potential to synthesize high-quality text, few studies exist that systematically investigate whether an LLM can effectively balance/improve FG under realistic settings surrounding fake news generation such as class imbalance and multilingual data. In addition, there exist abundant types of key augmentation settings (including the way they perform prompt strategy, amount of replacement and downsampling) that have not been heavily researched to find out if models really benefit from them.
Most recently, Chalehchaleh et al. \cite{chalehchaleh2025enhancing} addressed this deficiency by using Llama 3\cite{dubey2024llama} (an open source language learning model developed by Meta GenAI) for data augmentation on two multilingual datasets: TALLIP (which spans English, Hindi, Indonesian, Swahili and Vietnamese) and MM-COVID (which includes English as well as Spanish, French, Hindi, Italian and Portuguese). Chalehchaleh et al. \cite{chalehchaleh2025enhancing} extended their zero-shot method by exploring few-shot augmentation. They generated multiple synthetic samples per article. This allowed them to compare different augmentation amounts, sampling strategies, and class-wise approaches for fake news detection.
They found small but consistent improvements when using limited augmentation. Over-augmentation reduced performance by adding noise. They also noted that Bangla still lacks a systematic study, despite its class imbalance and unique linguistic challenges.

\subsection{Research Gap and Contribution}
Although the fake news detection research in English is relatively mature and preliminary datasets have been developed for Bangla but a gap exists in learning from modern LLM-based techniques for augmenting fake news detection performance for low-resource languages. The BanFakeNews dataset is useful as the first public Bangla fake news resource, but it can still be improved using systematic data augmentation. The class imbalance in the dataset (approximately 50,000 news articles with much more authentic than fake samples), together with the fact that both traditional machine learning and neural network techniques have trouble dealing with certain kinds of misinformation, justifies the use of LLM-based augmentation methods. It has been shown recently that LLM based augmentation can result in significant improvements with adequate setup, though such a result is yet to be verified for Bangla. A systematic analysis of how different augmentation strategies work in this context is necessary, especially because the linguistic nature of Bangla is different from those languages used in previous research and also keep in mind that a number of UGE datasets3 (including BanFakeNews) covers not only fake news but also clickbait, satire and misinformation. This paper addresses this gap by adopting effective augmentation strategies for Bangla fake news detection. 

\begin{figure*}[t]
\centering
\includegraphics[width=\textwidth]{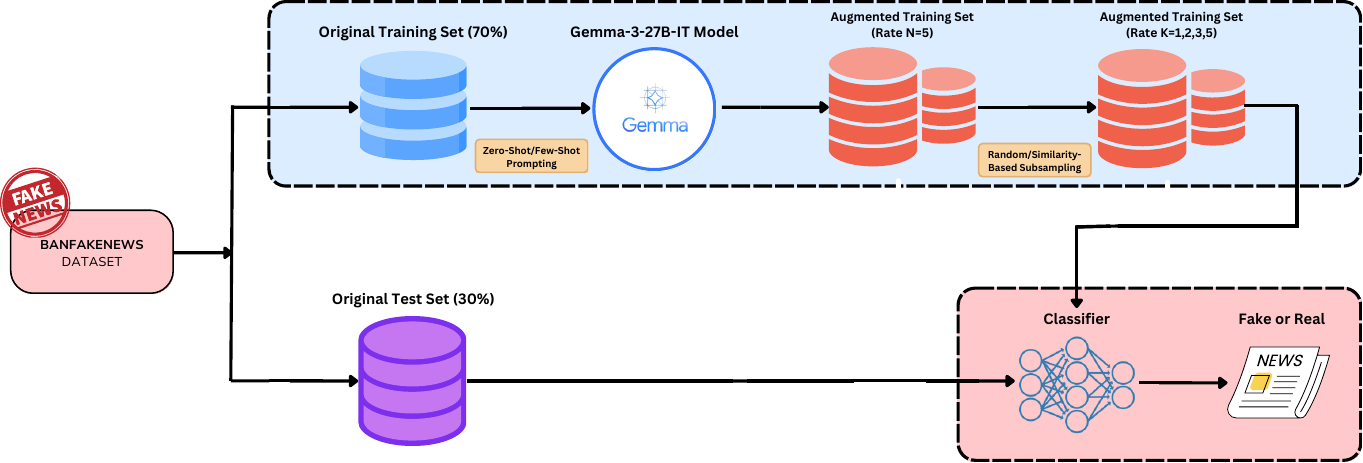}
\caption{Overview of the proposed methodology pipeline. The BanFakeNews dataset is split into training (70\%) and test (30\%) sets. The Gemma-3-27B-IT model generates $N=5$ synthetic paraphrases per original fake news article in the training set using zero-shot or few-shot prompting. From these candidates, $K$ samples (where $K \in \{1, 2, 3, 5\}$) are selected through either random or similarity-based subsampling to create the final augmented training set. The Bangla BERT classifier is then trained on the augmented set and evaluated on the original, non-augmented test set to predict fake or real news labels.}
\label{fig:methodology_pipeline}
\end{figure*}

\section{Proposed Methodology}
\label{sec:methodology}

To address the problem of scarcity of quality labeled data for Bangla fake news detection, we propose a method to generate synthetic news samples based on LLMs and augment the BanFakeNews dataset with them. The high-level view of our methodology is similar to the overall pipeline shown in \autoref{fig:methodology_pipeline}.

\subsection{LLM-Based Data Augmentation with Gemma}
LLMs for natural language generation have been demonstrated to be very powerful, having strong performance in the task of multilingual text generation and paraphrasing. In the current study, we use \textbf{Gemma 3}- an open-source LLM by Google DeepMind. Namely, we use the instruction-tuned \textbf{27B model} (\textbf{Gemma-3-27B-IT}) trained on dialogue and instruction-following data.
We choose Gemma-3-27B-IT based on (1) its open-source and budget friendly API key, (2) its decent performance on Natural Language Processing (NLP) based tasks \cite{abrar2024performance}, (3) large parameter size that ensures a reasonable trade-off between generation quality and computational tractability towards our task.

We investigate the augmentation regimes and adjust the number of parameters for Bangla fake news detection with zero-shot/few-shot prompting, varying augmentation rates through different subsampling methods, as well as class-specific oversampling schemes to address the data's intrinsic imbalance.

\subsubsection{Zero-Shot Versus Few-Shot Prompting}
\begin{figure*}[t]
    \centering
    \includegraphics[width=\textwidth]{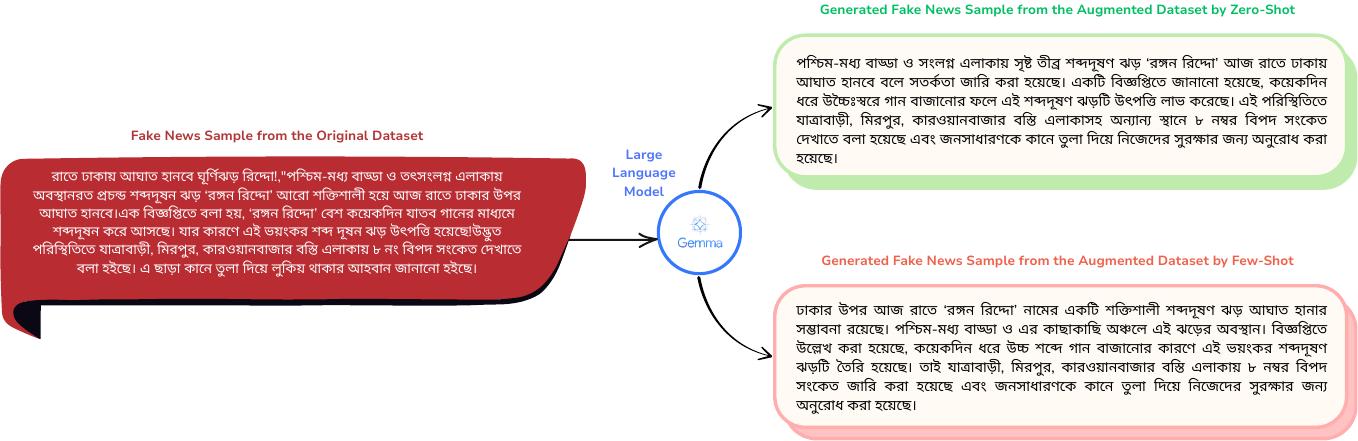}
    \caption{An example of LLM-based synthetic news generation for dataset augmentation.}
    \label{fig:dataset_augmentation}
\end{figure*}

\mbox{}\\
In LLM-augmentation, the prompt is essentially a gateway to summoning the desired generated output. For our purposes, prompts were designed thoroughly to prompt Gemma to generate several synthetic Bangla news articles while keeping strictly the semantic meaning and label (fake/real) of the former like \autoref{fig:dataset_augmentation}.

\textbf{Zero-shot prompting (ZS)} provides the model with no information other than natural language supervision and the target page, depending entirely on pre-trained knowledge to formulate $N=5$ paraphrases. Our designed prompt is given below:

\noindent\fbox{%
\parbox{0.97\linewidth}{%
\textbf{Prompt Template}

\vspace{0.5em}

Paraphrase the following news article delimited by triple backquotes in 5 different ways. The generated articles must retain the exact same meaning, facts, and label (real/fake) as the original article. Maintain the article format and length. Only return the articles and no extra explanation. Enclose each paraphrased article within [BEGINARTICLE] and [ENDARTICLE] tags.\\

``\{text\}"

\vspace{0.5em}
\textbf{NUMBERED GENERATED ARTICLES:}
}%
}

In \textbf{Few-shot prompting (FS)}, we give the LLM few examples directly in the prompt so that it can give us better result. In our case, we provided five examples with the prompt.

\subsubsection{Varying Augmentation Rates}
\mbox{}\\
The augmentation rate ($K$) is the number of augmented fake news article should be selected per original fake news article. To find the right trade-off between diversity, noise injection and model performance, our augmentation process is carried out in two stages:
\begin{enumerate}
 \item \textbf{Generation Stage:} For each piece of original fake news article in the training dataset, in both zero-shot and few-shot method, the LLM was asked to generate five synthetic variants ($N = 5$).
\item \textbf{Selection Stage:} In the selection stage, we choose $K$ samples from the $N$ generated ones to add to the final training set. 
We tested different values of $K$ \ $(K = 1, 2, 3, 5)$ to see how using fewer than five samples affects the results.
\end{enumerate}

\subsubsection{Selection Strategies: Random Versus Similarity-Based Subsampling}
\mbox{}\\
After generating and parsing N candidates, a greedy strategy is applied to retain the K best samples. All generated candidates are passed directly to the selection procedure.

\begin{enumerate}
\item \textbf{Similarity-Based Selection (S):} This strategy prefers samples that are most similar to the original text. Sentence embeddings are computed for both the original and generated texts using Sentence-BERT (SBERT), specifically the all-MiniLM-L6-v2 model.
The K samples with the highest cosine-similarity scores to their source article are selected.

\item \textbf{Random Selection (R):} This baseline strategy instead favors greater lexical diversity. K samples are selected uniformly at random from the entire pool of generated candidates (since no similarity-based filtering is applied).
\end{enumerate}

\subsubsection{Class-Specific Augmentation}
\mbox{}\\
To strategically address the notable class imbalance in the BanFakeNews dataset (approximately 5.5:1 ratio of real to fake news), we adopt a targeted \textbf{Fake-Only Augmentation} strategy. Instead of over-sampling both the classes or the majority class, we only deal with synthetic samples generation for minority Fake News class. This is consistent with the best result in imbalanced text classification and seems to produce both effective and stable performance improvements, as shown in our results.

\subsection{Dataset Preparation and Splitting Strategy}
The base dataset consists of 7,202 authentic (labeled) and 1,299 fake (labeled) news articles, totaling 8,501 samples. We split the dataset with a $\mathbf{70\%\text{--}30\%}$ train-test ratio, stratified by both label and category to maintain natural distributions.

The consistent non-augmented splits yielded:
\begin{itemize}
    \item \textbf{Original Train Split:} 5,041 Real articles and approximately 909 original Fake articles (70\% of 1,299).
    \item \textbf{Test Set (Original):} 2,551 articles (2,161 Real + 390 Fake). The test set remains completely non-augmented.
\end{itemize}

The training set was augmented exclusively in the Fake News class (Label 0), resulting in a variable total size shown in \autoref{tab:training_composition}.

\begin{table}[b]
\centering
\caption{Training Set Composition Across Augmentation Rates}
\label{tab:training_composition}
\begin{tabular*}{\columnwidth}{@{\extracolsep{\fill}}cccc@{}}
\toprule
\textbf{N/K Value} & \textbf{Total} & \textbf{Real} & \textbf{Fake} \\
\midrule
N = 5 & 10,495 & 5,041 & 5,454 \\
K = 3 & 8,677 & 5,041 & 3,636 \\
K = 2 & 7,768 & 5,041 & 2,727 \\
K = 1 & 6,859 & 5,041 & 1,818 \\
\bottomrule
\end{tabular*}
\end{table}

\subsection{BERT-Based Classification}
We used the \textbf{Bangla BERT Base model} sagorsarker/bangla-bert-base \cite{hossain2020banfakenews} for classification. This monolingual, domainadapted model has its advantage over generic multilingual models in recognizing fine-grained linguistic characteristics manifested in the Bangla language. We compute a linear classification layer on top of the final contextual embeddings to perform the binary classification (fake or real) and fine-tune end-to-end on our augmented training sets.

\section{Experiments}
\label{sec:experiments}

\subsection{Dataset Description and Statistics}
\begin{figure}[htbp]
    \centering
    \begin{tikzpicture}
        \begin{axis}[
            ybar,
            bar width=40pt,
            width=\columnwidth,
            height=6cm,
            ylabel={Number of Articles},
            symbolic x coords={Authentic News, Fake News},
            xtick=data,
            nodes near coords={\pgfmathprintnumber[precision=0]{\pgfplotspointmeta}},
            nodes near coords align={vertical},
            ymin=0,
            ymax=8000,
            ylabel style={font=\small},
            xlabel style={font=\small},
            tick label style={font=\small},
            x tick label style={align=center, font=\small},
            enlarge x limits=0.4,
            every node near coord/.append style={font=\footnotesize}
        ]
        \addplot[fill=blue!60] coordinates {(Authentic News,7202) (Fake News,1299)};
        \end{axis}
    \end{tikzpicture}
    \caption{BanFakeNews Dataset Statistics: Distribution of Authentic News (7,202 articles, 84.7\%) and Fake News (1,299 articles, 15.3\%). Total: 8,501 articles.}
    \label{fig:banfakenews_stats}
\end{figure}
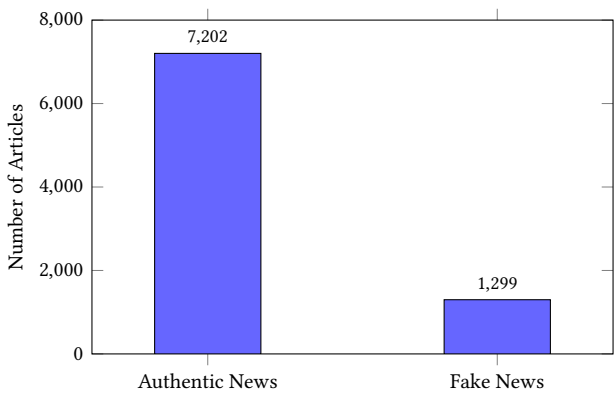

The BanFakeNews dataset \cite{hossain2020banfakenews} is a primary resource in the Bangla language, characterized by the statistics in \autoref{fig:banfakenews_stats}
 for the subsets used.

\begin{table*}
\caption{Performance Comparison Across Augmentation Configurations on the Test Set (R - Random Sub Sampling \& S - Similarity Based Sub Sampling)}
\label{tab:augmentation_results}
\centering
\small
\begin{tabular}{clcccccccc}
\toprule
\textbf{Result} & \textbf{Prompting} & \textbf{Configuration} & \textbf{Class} & \textbf{P} & \textbf{R} & \textbf{F1} & \textbf{Combined F1} & \textbf{Acc.} \\
\midrule
\textbf{1} & - & Baseline (No Aug.) & Fake (0) & 0.9104 & 0.8077 & 0.8560 & 0.9574 & 0.9584 \\
& & & Real (1) & 0.9660 & 0.9857 & 0.9757 & & \\
\midrule
\textbf{2} & Zero-Shot & \textbf{K=5 (Optimal}) & Fake (0) & 0.8857 & 0.8744 & \textbf{0.8800} & \textbf{0.9634} & \textbf{0.9635} \\
& & & Real (1) & 0.9774 & 0.9796 & \textbf{0.9785} & & \\
\midrule
\textbf{3} & Zero-Shot & R, K=3 & Fake (0) & 0.8579 & 0.8667 & 0.8622 & 0.9578 & 0.9577 \\
& & & Real (1) & 0.9759 & 0.9741 & 0.9750 & & \\
\midrule
\textbf{4} & Zero-Shot & S, K=1 & Fake (0) & 0.8904 & 0.8128 & 0.8499 & 0.9553 & 0.9561 \\
& & & Real (1) & 0.9667 & 0.9820 & 0.9743 & & \\
\midrule
\textbf{5} & Zero-Shot & S, K=2 & Fake (0) & 0.7422 & \textbf{0.9154} & 0.8197 & 0.9410 & 0.9385 \\
& & & Real (1) & \textbf{0.9841} & 0.9426 & 0.9629 & & \\
\midrule
\textbf{6} & Zero-Shot & S, K=3 & Fake (0) & 0.9104 & 0.8077 & 0.8560 & 0.9574 & 0.9584 \\
& & & Real (1) & 0.9660 & 0.9857 & 0.9757 & & \\
\midrule
\textbf{7} & Few-Shot & K=5 & Fake (0) & 0.8024 & 0.8538 & 0.8273 & 0.9462 & 0.9455 \\
& & & Real (1) & 0.9733 & 0.9621 & 0.9677 & & \\
\midrule
\textbf{8} & Few-Shot & R, K=3 & Fake (0) & 0.7968 & 0.9051 & 0.8475 & 0.9515 & 0.9502 \\
& & & Real (1) & 0.9824 & 0.9584 & 0.9703 & & \\
\midrule
\textbf{9} & Few-Shot & S, K=1 & Fake (0) & 0.9342 & 0.7282 & 0.8184 & 0.9480 & 0.9506 \\
& & & Real (1) & 0.9528 & 0.9907 & 0.9714 & & \\
\midrule
\textbf{10} & Few-Shot & S, K=2 & Fake (0) & 0.7797 & 0.9077 & 0.8389 & 0.9483 & 0.9467 \\
& & & Real (1) & 0.9828 & 0.9537 & 0.9681 & & \\
\midrule
\textbf{11} & Few-Shot & S, K=3 & Fake (0) & \textbf{0.9664} & 0.7385 & 0.8372 & 0.9536 & 0.9561 \\
& & & Real (1) & 0.9547 & \textbf{0.9954} & 0.9746 & & \\
\bottomrule
\end{tabular}
\end{table*} 

\subsection{Experimental Settings}

\subsubsection{Model Configurations and Infrastructure}
\mbox{}\\
All LLM generation was performed using the $\mathbf{4\text{-}bit}$ quantized, instruction-tuned \texttt{Gemma-3-27B-IT} model. Key generation parameters included:
\begin{itemize}
    \item $N=5$ generations per article.
    \item Temperature: Default 1.
    \item The downstream classification model, \\ \texttt{sagorsarker/bangla-bert-base}, was fine-tuned for 3 epochs using the AdamW optimizer with a learning rate of $2 \times 10^{-5}$ and a batch size of 8.
    \item SBERT (\texttt{all-MiniLM-L6-v2}) was used for similarity scoring.
\end{itemize}

\subsubsection{Evaluation Metrics}
\mbox{}\\
To comprehensively assess model performance, especially given the class imbalance, we utilize Precision, Recall, and the F1 score, reported explicitly for the minority Fake News class (Label 0).
\begin{equation}
\text{Precision} = \frac{\text{TP}}{\text{TP} + \text{FP}}
\end{equation}

\begin{equation}
\text{Recall} = \frac{\text{TP}}{\text{TP} + \text{FN}}
\end{equation}

\begin{equation}
\text{F1} =
2 \times \frac{\text{Precision} \times \text{Recall}}{\text{Precision} + \text{Recall}}
\end{equation}

\section{Results and Analysis}
\label{sec:results}

In this section, we present and analyze the experimental results, focusing on the F1 score of the minority Fake News class (Label 0) on the original, non-augmented test set.

\subsection{Baseline and Augmentation Performance Summary}
The performance baseline, achieved without any augmentation, yielded a Fake News F1 score of $\mathbf{0.8560}$ (\autoref{tab:augmentation_results}, Result 1). The optimal configuration was determined to be \textbf{Zero-Shot prompting with Random Subsampling at $K=5$} (Result 2), resulting in a Fake News F1 score of $\mathbf{0.8800}$, marking an improvement of $\mathbf{2.4}$ points over the baseline.

\subsection{Impact of Class-Specific Augmentation}
In all cases the analysis shows that augmenting \textit{only the fake news class} is the better strategy. Augmented both classes or real class only gave marginal improvement or even declining performance against the base line, suggesting that targeted minority class oversampling is more suitable for overcoming the class imbalance.

\subsection{Effect of Augmentation Rate}
Our experiments revealed a complex relationship between the augmentation rate ($K$) and model performance. The highest F1 score ($\mathbf{0.8800}$) was achieved at the highest rate of $\mathbf{K=5}$ (ZS/R). The relationship between augmentation rate and performance varied significantly across different subsampling strategies:
\begin{itemize}
    \item With random subsampling (ZS/R), performance at $K=3$ (Result 3) reached $0.8622$ F1, and increased further to the optimal $0.8800$ F1 at $K=5$ (Result 2).
    \item With similarity-based subsampling, the pattern differed: $K=1$ (ZS/S, Result 4) achieved $0.8499$ F1, while $K=2$ (Result 5) dropped significantly to $0.8197$ F1, and $K=3$ (Result 6) matched the baseline at $0.8560$ F1.
\end{itemize}
This highlights that higher augmentation rates can be beneficial when combined with random subsampling, which effectively leverages diversity, whereas similarity-based selection shows more inconsistent behavior across different rates.

\subsection{Prompting Strategies: Zero-Shot versus Few-Shot}
The \textbf{Zero-Shot (ZS)} approach consistently and significantly outperformed Few-Shot (FS) prompting across all comparable configurations. The ZS/R/$K=5$ configuration achieved the maximum $\mathbf{0.8800}$ F1. In contrast, all tested Few-Shot configurations performed at or below the non-augmented baseline (F1 $\le 0.8560$), with the best few-shot result being $0.8475$ F1 (FS/R/$K=3$, Result 8). This confirms that ZS effectively leveraged the LLM's full internal knowledge, resulting in greater diversity and a more effective training signal for the classifier.

\subsection{Subsampling Strategies: Random versus Similarity-Based}
\textit{Random Subsampling (R)} was the most effective strategy for exceeding baseline performance with zero-shot prompting.
\begin{itemize}
    \item At $K=5$, Random Subsampling (ZS/R, Result 2) achieved the optimal $0.8800$ F1, representing the best overall performance.
    \item At $K=3$, Random Subsampling (ZS/R, Result 3) also showed strong performance at $0.8622$ F1.
    \item In contrast, Similarity-based selection showed inconsistent results: $K=1$ (ZS/S, Result 4) achieved $0.8499$ F1, $K=2$ (Result 5) dropped to $0.8197$ F1, and $K=3$ (Result 6) only matched the baseline at $0.8560$ F1.
\end{itemize}
This conclusive evidence shows that prioritizing linguistic \textbf{diversity} (Random Sampling) over strict semantic \textbf{similarity} (SBERT Sampling) is essential for maximizing performance with zero-shot prompting.

\section{Discussion}
\label{sec:discussion}

The experimental results demonstrate that, provided with cautious control of the approach, LLM-based data augmentation is an effective strategy for improving Bangla fake news detection. Applying Zero-Shot Random Subsampling to only the fake class at a high augmentation rate of $K$=$5$ produced the largest F1 gain of $\mathbf{2.4}$ points, which effectively addressed the class imbalance issue.

The most important observation is the critical interaction between augmentation rate and subsampling strategy. The maximum improvement of $2.4$ points at $K$=$5$ with random subsampling demonstrates that the classifier can benefit substantially from higher volumes of diverse synthetic data. Interestingly, the performance improvement is progressive with random subsampling: $K=3$ achieved $0.8622$ F1 (improvement of $0.62$ points), and $K=5$ reached $0.8800$ F1 (improvement of $2.4$ points). However, this positive relationship holds specifically for random subsampling. With similarity-based selection, higher rates did not yield consistent improvements, with $K=2$ showing particularly poor performance ($0.8197$ F1), suggesting that selecting semantically similar samples may lead to reduced diversity and potentially introduce repetitive patterns that hinder generalization.

In addition, the Zero-Shot method is better than Few-Shot, contrary to literature on LLMs in general which tends to favour Few-Shot. In this particular, low-resource Bangla setting the ZS prompt seemed to provide a more beneficial diversity of paraphrases. This result is important for practical use as ZS prompting is a much less effort (or engineering) intensive process than the careful curation of high quality FS examples.

Lastly, the superior performance of Random Subsampling compared to strict Similarity-based selection (when applied post-filtering) demonstrates the significance of diversity in sampling signals. The combination of a tight quality floor ($\text{cos} > 0.7$) followed by random selection at higher augmentation rates provides a highly efficient procedure for creating generalization-enhancing artificial training data. The success of the $K=5$ configuration suggests that, when diversity is preserved through random sampling, the classifier can effectively learn from a larger volume of synthetic data without overfitting to artificial patterns.

\section{Conclusion}
\label{sec:conclusion}

This study addressed the scarcity and imbalance of Bangla fake news data by introducing an LLM-based augmentation framework tailored to the BanFakeNews dataset. Our experiments show that selectively augmenting only the minority fake news class at a high augmentation rate (K = 5), together with random subsampling, produces the strongest performance gains, with the best configuration—Zero-Shot prompting with Random Subsampling—improving the Fake News F1 score to 0.8800, a 2.4-point increase over the baseline. These findings highlight the potential of LLM-driven augmentation for strengthening fake news detection systems in low-resource languages, particularly when diversity-preserving sampling strategies are used. While the approach demonstrates clear benefits, it remains sensitive to prompt design, augmentation rate, and subsampling strategy, indicating the need for more granular evaluation with additional Bangla-specific encoders and cross-dataset validation. Future work may explore broader multilingual comparisons, adaptive prompt engineering, and integration of multimodal cues to further improve robustness. Overall, our results underscore that carefully controlled LLM-based augmentation can be a practical and effective solution for enhancing fake news detection in under-resourced linguistic settings.

\bibliographystyle{ACM-Reference-Format}
\bibliography{references}
\end{document}